\documentclass{article}

\usepackage{arxiv}

\usepackage[utf8]{inputenc} 
\usepackage[T1]{fontenc}    
\usepackage{xcolor}         
\usepackage{hyperref}       
\usepackage{url}            
\usepackage{booktabs}       
\usepackage{amsmath}        
\usepackage{amssymb}
\usepackage{graphicx}
\usepackage{adjustbox}      

\hypersetup{
  colorlinks=true, linkcolor=blue, citecolor=blue, urlcolor=blue,
  pdftitle={Humanoid Loco-Manipulation With Discrete VLA Model},
  pdfauthor={Wenxin Shao, Siqi Chai, Kun Li, Kerou Zhang, Xinzhou Jiang, Wei Xu, Qiang Liu}
}

\title{Humanoid Loco-Manipulation With\\Discrete VLA Model}
\renewcommand{\shorttitle}{Humanoid Loco-Manipulation With Discrete VLA Model}
\renewcommand{\headeright}{}   
\renewcommand{\undertitle}{\mbox{}} 
\date{}

\author{
  Wenxin Shao$^{*}$\quad Siqi Chai$^{*}$\quad Kun Li$^{*}$\quad Kerou Zhang\quad Xinzhou Jiang\quad Wei Xu\quad Qiang Liu$^{\dagger}$ \\[4pt]
  \normalfont Horizon Robotics \\[3pt]
  \normalfont\small $^{*}$Equal contribution\quad $^{\dagger}$Corresponding author \\[3pt]
  \normalfont\small Project page: \url{https://horizonrobotics.github.io/gail/Holo-M/}
}

\begin{document}

\maketitle

\begin{abstract}

Vision-language-action (VLA) models using
discrete action tokens have proven effective for
controling robotic arms on manipulation tasks.
For a humanoid, however, the whole-body
action space -- legs, torso, arms, and hands -- is far higher-dimensional and
heterogeneous, raising tokenization, training, and real-time inference
challenges that the previous VLA models do not address. We present
Holo-M, to our knowledge the first discrete VLA model for humanoid
loco-manipulation that intrinsically exploits the language model by
extending its vocabulary with action tokens. In this model, we
devise a unified action tokenizer that decomposes the humanoid action
space into four body-part-specific tokenizers -- end-effector, body, hand, and
kinematics -- enabling training across drastically different
embodiments and data sources, including humanoid teleoperation, ego-centric
human video, and simulation. By extending the language model's vocabulary with these action tokens, we avoid the knowledge-insulation problem inherent to the models that use separate
continuous action experts. To meet real-time control requirements, we decode
each body part's action tokens through grouped discrete diffusion decoding,
rather than using autoregression on the action tokens. We have conducted extensive experiments on the SIMPLE humanoid loco-manipulation benchmark, in which
Holo-M achieves the highest
success rates in both the generalist and specialist evaluations, leading the second best by significant margins. We will release all the code and model weights.

\end{abstract}

\section{Introduction}
\label{sec:intro}

Humanoid robots are, in principle, the most general embodiment for operating
in human environments, since a single whole-body morphology can both walk
and manipulate. Realizing this requires \emph{loco-manipulation} policies
that couple locomotion and manipulation under open-ended, language-specified
instructions. Vision-language-action (VLA) models, which fine-tune
vision-language backbones to emit robot actions, have emerged as a leading
recipe for such generalist policies.

A key design choice in VLA models is how actions are represented to the
underlying language model. One line of work represents actions as
\emph{discrete tokens} drawn from the same vocabulary as text, so action
generation becomes another sequence-modeling problem the pretrained backbone
already knows how to solve~\cite{rt2,openvla}. This has so far been developed
almost exclusively for \emph{robotic arms}, whose low-dimensional action
space (a handful of joint angles or an end-effector pose plus a gripper)
lets a single tokenizer and either an autoregressive or, more recently,
discrete diffusion ~\cite{discretediffusionvla} tokenize, train, and
decode efficiently. Humanoids have not received the same recipe -- existing
humanoid VLA models instead rely on \emph{continuous} action generation,
typically a flow-matching or diffusion action expert attached to a VLM
backbone~\cite{psi0}.

This gap is not incidental. A humanoid's whole-body action space -- legs,
torso, arms, and dexterous hands -- has far more degrees of freedom than a
single arm or wheeled bimanual manipulator, and is heterogeneous across very
different scales and timescales. Naively extending arm-style solutions runs
into three compounding problems: \emph{tokenization}, since a flat
vocabulary is either too coarse or too large to fit a usable context;
\emph{training}, since no single dataset covers all of a humanoid's DoF,
requiring learning from drastically different, partially overlapping
sources; and \emph{real-time inference}, since autoregressive decoding over
a much longer token sequence is too slow for whole-body control.

We address tokenization and training with a \emph{generic, per-body-part
tokenization space}: rather than one monolithic tokenizer, we decompose the
action space into four parts -- end-effector (EEF: wrist and fingertip
pose), body, hand, and kinematics -- each with its own dedicated tokenizer.
This lets drastically different, otherwise incompatible embodiments and
data sources -- teleoperated humanoid, large-scale human-centric data, and
simulation -- each supervise only the body parts they cover, rather than
requiring every source to provide full whole-body ground truth.

We address real-time inference with \emph{grouped discrete diffusion}: instead of
decoding tokens autoregressively, our model generates each body part's
tokens in parallel through a small, fixed number of de-masking steps, so
wall-clock cost no longer grows with the considerably longer action-token
sequence a humanoid requires.

Keeping actions as tokens in the same vocabulary as language, rather than
routing them through a separate continuous action expert, is also
architecturally advantageous. Continuous action experts must be carefully
gradient-insulated from the pretrained VLM backbone to avoid corrupting its
semantic representations~\cite{knowledgeinsulation}; our model has no such
boundary to insulate, since action and language tokens are produced by the
same sequence model over the same vocabulary.

\begin{figure}[t]
   \centering
   \includegraphics[width=0.55\columnwidth]{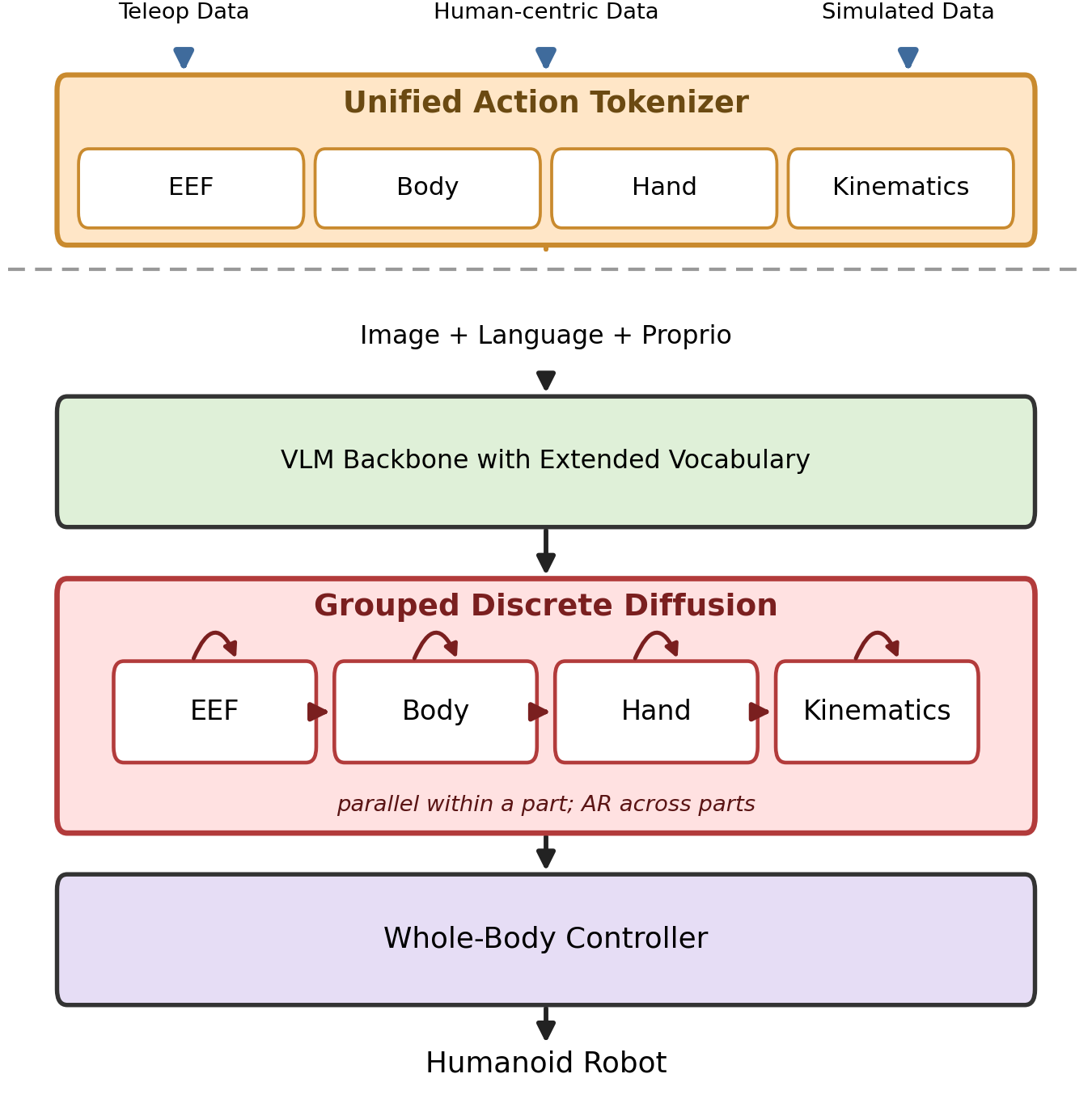}
   \caption{Overview of the proposed Holo-M framework. A unified action
     tokenizer, pre-trained across heterogeneous data (teleoperated,
     human-centric, simulated), decomposes the humanoid action space
     into four body-part tokenizers (EEF, body, hand, kinematics) sharing
     one discrete vocabulary. At inference, the VLM backbone encodes the
     image, language, and proprioceptive state into this vocabulary and
     decodes action tokens via grouped discrete diffusion -- parallel within a body
     part, autoregressive across parts -- which are then detokenized into
     whole-body controller commands driving the humanoid.}
   \label{fig:overview}
\end{figure}

As shown in Fig.~\ref{fig:overview}, we make the following
contributions:
\begin{enumerate}
  \item A \textbf{unified action tokenizer} for humanoids, decomposed into four
    body-part-specific tokenizers (EEF, body, hand, kinematics), that unifies
    different embodiments and data sources -- teleoperated humanoid, human-centric video, and simulation -- within a single
    discrete action vocabulary.
  \item \textbf{Holo-M}, to the best of our knowledge the \textbf{first discrete VLA model for humanoid loco-manipulation that intrinsically exploits the language model} by extending its vocabulary with action tokens, letting action generation benefit directly from the backbone's ability.
  \item A \textbf{grouped discrete diffusion} decoding scheme -- parallel
    within each body part, autoregressive across parts -- making real-time
    inference tractable for a humanoid's action space.
\end{enumerate}
We validate these contributions on the SIMPLE humanoid loco-manipulation
benchmark~\cite{simple}, where our model outperforms all the
baseline models ~\cite{psi0} achieving the highest success
rate in both the generalist and specialist evaluation modes.

\section{Related Work}
\label{sec:related}

\subsection{Vision-Language-Action Models and Action Decoding}
VLA models differ mainly in how actions are decoded. \emph{Continuous}
decoders attach a diffusion or flow-matching action expert to the VLM
backbone (e.g., $\pi_0$~\cite{pi0} for arms, $\Psi_0$~\cite{psi0} for
humanoids), requiring a separate, non-token module. \emph{Discrete}
decoders instead represent actions as language-vocabulary tokens,
decoded autoregressively~\cite{rt2,openvla} or via discrete
diffusion~\cite{discretediffusionvla}; \cite{g05} similarly unifies
reasoning and action autoregressively for arm/mobile-manipulation
robots. This unification has so far only been built for robotic arms.
Concurrent humanoid work, WholeBodyVLA~\cite{wholebodyvla}, predicts
discrete VQ codes for manipulation and locomotion separately, but
grounds them via a separate execution decoder and downstream RL
controller rather than sharing the VLM's vocabulary. We instead keep body-part tokens in the same
vocabulary and sequence model as language, decoded end-to-end via
grouped discrete diffusion, for a humanoid's far larger, heterogeneous
action space.

\subsection{Action Tokenization for Robot Learning}
Effective tokenization is central to discrete VLA models. Early schemes
uniformly bin each action dimension per timestep, which scales poorly to
high-frequency, dexterous actions; FAST~\cite{fast} instead compresses
action chunks via the discrete cosine transform, and FASTer~\cite{faster}
replaces this with a learned vector-quantized tokenizer and block-wise
autoregressive decoding -- but both target one flat vocabulary sized for
arm-scale action spaces. Since a humanoid's action space spans locomotion,
posture, and dexterous manipulation at very different scales, we instead
learn four separate, body-part-specific tokenizers, each specialized to its
own action statistics and data source.

\subsection{Humanoid Whole-Body Control}
A separate line of work develops low-level whole-body controllers (WBCs)
that track high-level commands while maintaining balance, sitting beneath
rather than replacing a high-level policy: HoloMotion~\cite{holomotion}
learns a generalist motion-tracking policy from a large-scale hybrid motion
corpus, SONIC~\cite{sonic} scales motion tracking to a single unified
controller, and decoupled WBC~\cite{decoupledwbc} separates a
reinforcement-learned lower body from an IK-driven upper body, as used by
the $\Psi_0$ baseline~\cite{psi0}, which we also adopt so that our
comparison isolates the high-level policy's effect rather than the
controller's.

\subsection{Discrete Diffusion}
Discrete diffusion decoding iteratively unmasks tokens in parallel rather
than one at a time, originally proposed for image generation~\cite{maskgit}
and recently adapted to robots, where an entire action chunk is de-masked
within a single transformer pass~\cite{discretediffusionvla}. For a
humanoid we instead use a hierarchical schedule -- parallel \emph{within}
each body part but autoregressive \emph{across} parts, so later parts
condition on already-decoded ones -- retaining most of the speed advantage
over full autoregression while respecting inter-part dependencies.

\section{Method}
\label{sec:method}

\subsection{Unified Action Tokenizer for Humanoids}
\label{sec:action_tokenizer}
Designing a single action tokenizer for humanoid loco-manipulation is
challenging along two axes. First, teleoperation logs, egocentric human
video, and simulation each expose only a subset of the full humanoid
state, so a tokenizer built around any one source's native
representation cannot ingest the others without discarding information
or resorting to artificial padding. Second, human motion is naturally
described by biomechanical degrees of freedom, whereas humanoid control
may be expressed through both a kinematic joint chain and task-space
end-effector targets; a unified tokenizer must therefore bridge
biomechanics and robot kinematics rather than treat one as a subset of
the other. We address both by factorizing the action space into four
independently tokenized groups, with the EEF group giving a shared
task-space representation for training only, and per-sample missing groups excluded from
the loss rather than filled with artificial targets.

\begin{table}[t]
\centering
\caption{Canonical action-space decomposition used by Holo-M.}
\label{tab:canonical_action_space}
\small
\setlength{\tabcolsep}{5pt}
\begin{tabular}{lccc}
\toprule
Group & Representation & Dim. & Tokens \\
\midrule
End effector & Wrist and fingertip poses & 48 & 100 \\
Body & Body joint targets & 29 & 62 \\
Hand & Hand joint targets & 14 & 32 \\
Kinematics & Base-motion commands & 5 & 14 \\
\midrule
Full action & All groups & 96 & 208 \\
\bottomrule
\end{tabular}
\end{table}

Table~\ref{tab:canonical_action_space} defines the canonical action
representation used throughout Holo-M. Each action chunk contains
$T=30$ steps at 30\,Hz and follows the fixed group order
EEF $\rightarrow$ body $\rightarrow$ hand $\rightarrow$ kinematics:
\begin{equation}
A_t =
\left[
A_t^{\mathrm{eef}}
\,\middle\vert\,
A_t^{\mathrm{body}}
\,\middle\vert\,
A_t^{\mathrm{hand}}
\,\middle\vert\,
A_t^{\mathrm{kinematics}}
\right]
\in \mathbb{R}^{30\times96}.
\label{eq:canonical_action}
\end{equation}

The corresponding group dimensions are
\begin{equation}
\left(
D_{\mathrm{eef}},
D_{\mathrm{body}},
D_{\mathrm{hand}},
D_{\mathrm{kinematics}}
\right)
=
(48,29,14,5).
\label{eq:action_group_dims}
\end{equation}

The EEF group contains bimanual wrist positions, 6D wrist rotations,
and fingertip positions in the camera frame. The body and hand groups
contain joint-space targets in fixed hardware-specific orders. The
kinematics group contains base velocity, height, and yaw-rate commands. 

\paragraph{FAST-RVQ tokenization}
We introduce FAST-RVQ to map each continuous action group to a
fixed-length token sequence. Let
\[
\mathcal{G}
=
(\mathrm{eef},\mathrm{body},\mathrm{hand},\mathrm{kinematics})
\]
denote the ordered action groups. For group $g\in\mathcal{G}$, the
corresponding action chunk is
\begin{equation}
X_g =
\begin{bmatrix}
a_{t,g} & a_{t+1,g} & \cdots & a_{t+T-1,g}
\end{bmatrix}^{\top}
\in \mathbb{R}^{T\times D_g}.
\label{eq:group_action_chunk}
\end{equation}

Following FAST~\cite{fast}, we apply the discrete cosine transform
(DCT) along the temporal axis of each action dimension. FAST applies
byte-pair encoding to quantized DCT coefficients, which produces
variable-length token sequences. Such sequences are convenient for
autoregressive generation but less suitable for masked parallel
decoding, which benefits from fixed group boundaries.

FAST-RVQ instead retains the first $N$ temporal DCT coefficients and
uses residual vector quantization (RVQ) to encode the remaining
time-domain residual. We define the coefficient projection matrix as
\begin{equation}
P_N =
\operatorname{diag}\!\left(
\underbrace{1,\ldots,1}_{N},
\underbrace{0,\ldots,0}_{T-N}
\right)
\in \mathbb{R}^{T\times T}.
\label{eq:fast_rvq_projection}
\end{equation}
For each group, the low-frequency reconstruction and its residual are
\begin{equation}
\begin{aligned}
\widehat{X}_{g,N}
&=
\operatorname{IDCT}\!\left(
P_N\operatorname{DCT}(X_g)
\right),\\
R_{g,N}
&=
X_g-\widehat{X}_{g,N},
\end{aligned}
\label{eq:fast_rvq_residual}
\end{equation}
where DCT and IDCT are applied independently along the temporal axis.

We apply RVQ independently to $R_{g,N}$ for each action group. Each RVQ
stage uses a codebook fitted by $k$-means, without training a separate
neural tokenizer. In our implementation, we retain $N=2$ DCT
coefficients per action dimension and use $Q=4$ RVQ stages per group.
The token length of group $g$ is therefore
\begin{equation}
L_g = ND_g + Q = 2D_g+4.
\label{eq:group_token_count}
\end{equation}
Following the dimensions in
Table~\ref{tab:canonical_action_space}, the four groups contain
\begin{equation}
\left(
L_{\mathrm{eef}},
L_{\mathrm{body}},
L_{\mathrm{hand}},
L_{\mathrm{kin}}
\right)
=
(100,62,32,14)
\label{eq:group_token_lengths}
\end{equation}
tokens, respectively. The complete action sequence contains
\begin{equation}
L_{\mathrm{action}}
=
\sum_{g\in\mathcal{G}}L_g
=
208
\label{eq:fast_rvq_token_count}
\end{equation}
action tokens. Structural tokens marking group boundaries are not
included in this count. With the above settings, FAST-RVQ achieves a reconstruction error of 0.0068 on our validation action chunks from the combined datasets (Sec. \ref{sec:datasets_benchmark}) used in our experiments. 

\subsection{VLM Backbone with Discrete Action Tokens}
We use Qwen3-VL-2B~\cite{qwen3vl} as the vision-language
backbone. For robot demonstrations, the visual input is the current
RGB image from the head-mounted camera. For human demonstrations, the
model receives the corresponding egocentric RGB image. Robot images
are resized to $640\times360$ pixels and processed using the pretrained
Qwen3-VL image processor and vision encoder. During training, we apply
color jitter to brightness, contrast, saturation, and hue. The
resulting visual embeddings are placed in the same transformer context
as the language, setup, state, and action tokens.

\paragraph{Conditioning context}
The conditioning context follows the fixed order
image $\rightarrow$ task $\rightarrow$ setup $\rightarrow$ state:
\begin{equation}
c_t =
\left[
v_t
\,\middle\Vert\,
q^{\mathrm{task}}
\,\middle\Vert\,
q^{\mathrm{setup}}
\,\middle\Vert\,
q_t^{\mathrm{state}}
\right],
\label{eq:conditioning_context}
\end{equation}
where $v_t$ denotes the visual embeddings and $\Vert$ denotes sequence
concatenation. The task instruction begins with \texttt{Task:}. The
setup field identifies the embodiment and data source and is enclosed
by \texttt{<setup\_start>} and \texttt{<setup\_end>}. Examples include
``G1 humanoid robot from Humanoid Everyday'' and
``Human egocentric data from EgoDex.''
The state field contains selected robot joint angles in a fixed order.
Each angle is clipped to $[-\pi,\pi]$, normalized by $\pi$, and
uniformly quantized into one of 256 state-bin tokens. The tokens are
enclosed by \texttt{<state\_start>} and
\texttt{<state\_end>}. State
tokens provide conditioning information and are not included in the
208-token action sequence.

The conditioning prompt has the following structure:

\begingroup
\scriptsize
\begin{verbatim}
[image embeddings]
Task: [language instruction]
<setup_start>[embodiment and data source]<setup_end>
<state_start>[quantized state tokens]<state_end>
\end{verbatim}
\endgroup

\paragraph{Action-token sequence}
The unified tokenizer in Sec.~\ref{sec:action_tokenizer} maps each
action chunk to the canonical token sequence $y_t$. We append this
sequence to the conditioning context and train the backbone to model
\begin{equation}
p_\theta(y_t\mid c_t)
=
\prod_{g\in\mathcal{G}}
\prod_{j=1}^{L_g}
p_\theta
\left(
y_{t,j}^{g}
\mid
c_t,y_t^{<g},y_{t,<j}^{g}
\right),
\label{eq:grouped_ar_factorization}
\end{equation}
where $\mathcal{G}$, $L_g$, and the canonical group order are defined
in Sec.~\ref{sec:action_tokenizer}. Here, $y_t^{<g}$ denotes the tokens
from preceding groups, and $y_{t,<j}^{g}$ denotes the preceding tokens
within group $g$.

Action groups unavailable in a training sample are excluded from the loss
using the group-validity mask defined in
Sec.~\ref{sec:action_tokenizer}. We extend the Qwen3-VL vocabulary with
state-bin, action, and structural tokens. The action sequence has the following structure:

\begingroup
\scriptsize
\begin{verbatim}
Action: <action_output><action_start>
  <end_effector_start>[100 tokens]<end_effector_end>
  <body_start>[62 tokens]<body_end>
  <hand_start>[32 tokens]<hand_end>
  <kinematics_start>[14 tokens]<kinematics_end>
<action_end>
\end{verbatim}
\endgroup

\subsection{Grouped Discrete Diffusion for Efficient Action Decoding}
\label{sec:grouped_diffusion}

Autoregressive decoding requires one model evaluation for each action
token. For Holo-M, this results in 208 sequential evaluations per
action chunk. We reduce this latency by fine-tuning the autoregressive
policy with a grouped masked-diffusion objective based on
DiffusionVL~\cite{diffusionvl}. This adaptation retains the same VLM
backbone, action vocabulary, and canonical action representation.

\paragraph{Grouped masked-token training}
We use the four action groups defined in
Table~\ref{tab:canonical_action_space}, following the fixed order
EEF $\rightarrow$ body $\rightarrow$ hand $\rightarrow$ kinematics.
Each group forms one diffusion block. Tokens interact bidirectionally
within the current block, allowing them to be reconstructed in
parallel. Dependencies across blocks remain causal, so the current
group can attend to the conditioning context and all preceding groups,
but not to future groups.

Let $y_t^g=(y_{t,1}^g,\ldots,y_{t,L_g}^g)$ denote the clean token
sequence for group $g$. For each valid group, we independently sample
a masking probability $p_g$ and mask the corresponding subset of valid
token positions $\mathcal{M}_g$. Let $N_g$ denote the number of valid
tokens in group $g$ for the current sample. The corrupted sequence is
\begin{equation}
\tilde{y}_{t,j}^g =
\begin{cases}
\texttt{[MASK]}, & j\in\mathcal{M}_g,\\
y_{t,j}^g,       & j\notin\mathcal{M}_g.
\end{cases}
\label{eq:group_corruption}
\end{equation}
The model predicts the masked tokens conditioned on the multimodal
context, preceding action groups, and visible tokens in the current
group:
\begin{equation}
\mathcal{L}_g =
-\frac{1}{p_g N_g}
\sum_{j\in\mathcal{M}_g}
\log p_\theta
\left(
y_{t,j}^g
\mid
c_t,y_t^{<g},\tilde{y}_t^g
\right),
\label{eq:group_diffusion_loss}
\end{equation}
where $c_t$ is the conditioning context defined in
Eq.~\eqref{eq:conditioning_context}, $y_t^{<g}$ contains the tokens
from all preceding action groups, and $p_gN_g$ is the expected number
of masked valid tokens in group $g$.

Some data sources provide supervision for only a subset of the action
groups. Let $v_g\in\{0,1\}$ indicate whether group $g$ is available in
the current sample. The full objective is
\begin{equation}
\mathcal{L}_{\mathrm{diff}}
=
\frac{
\sum_{g\in\mathcal{G}} w_g v_g \mathcal{L}_g
}{
\sum_{g\in\mathcal{G}} w_g v_g
},
\label{eq:diff_loss}
\end{equation}
where $w_g$ controls the relative contribution of each action group.
Every training sample contains at least one valid action group. This
objective updates the post-trained VLM directly and does not introduce
a separate diffusion model or action head.

\paragraph{Confidence-based de-masking}
At inference, the four groups are decoded sequentially in their
canonical order, while tokens within each group are generated in
parallel. All positions in the current group are initialized with
\texttt{[MASK]}. At each iteration, the model predicts every masked
position and assigns each proposal a confidence score based on its
predicted probability. Positions with higher-confidence are demasked, while lower-confidence positions remain masked for the next iteration.
The number of demasked positions each iteration is decided by action token length of the current group and number of diffusion iterations.
Once a group is complete, its tokens are appended to the causal
context used to decode the next group. The conditioning prefix and
completed groups are cached across iterations.

This grouped
de-masking substantially reduces the sequential decoding depth. For
example, with eight de-masking iterations per group, decoding the four
action groups requires only 32 sequential decoding steps, compared
with 208 steps for token-by-token autoregressive decoding, yielding a
$6.5\times$ reduction in sequential decoding depth.

\subsection{Training Pipeline}
\label{sec:training_pipeline}

\begin{figure*}[t]
    \centering
    \includegraphics[width=\textwidth]
    {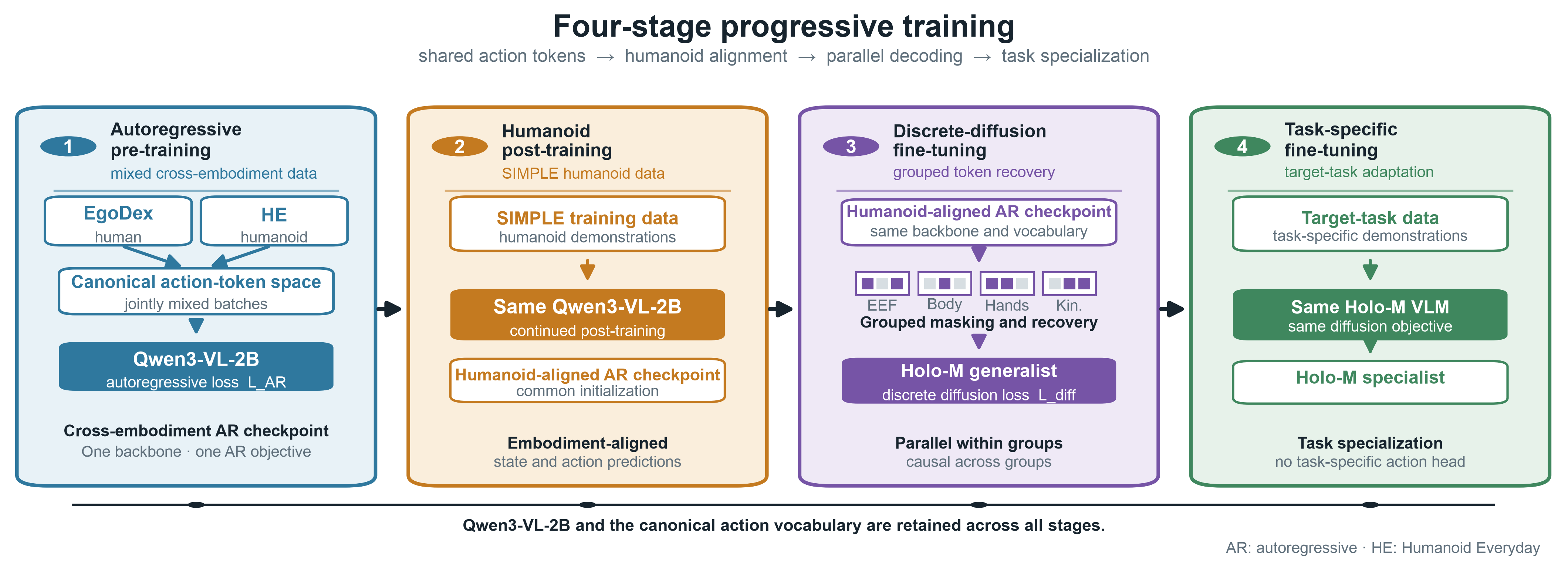}
    \caption{Progressive training of Holo-M through
    cross-embodiment autoregressive pre-training, humanoid
    post-training, grouped discrete diffusion fine-tuning, and
    task-specific adaptation.}
    \label{fig:training_pipeline}
\end{figure*}

Figure~\ref{fig:training_pipeline} summarizes the four-stage training
pipeline. The same Qwen3-VL-2B backbone and canonical action vocabulary
are retained throughout all stages. The pipeline first learns from
mixed human and humanoid demonstrations, aligns the policy with the
target humanoid embodiment, and then replaces token-by-token generation
with grouped de-masking. The final stage optionally specializes this
generalist policy using target-task demonstrations. No separate action
model or task-specific action head is introduced. The vision encoder is frozen, while the language backbone is optimized for action-token prediction.

\paragraph{Stage I: autoregressive pre-training}

We jointly pre-train the model on EgoDex~\cite{egodex} and Humanoid
Everyday~\cite{humanoideveryday}. EgoDex provides diverse egocentric
human manipulation, while Humanoid Everyday provides observations and
actions from a humanoid embodiment. The canonical action representation
allows both datasets to be mixed within the same training stage.
Each sample supervises only its available action groups through the
same group-validity mask $v_g$ used in
Eq.~\eqref{eq:diff_loss}: for example, an EgoDex sample has
$v_{\mathrm{eef}}=1$ and $v_{\mathrm{body}}=v_{\mathrm{hand}}=v_{\mathrm{kin}}=0$,
so only its EEF tokens contribute to the loss. The model is optimized
using the autoregressive objective in Eq.~\eqref{eq:ar_loss}.

\begin{equation}
\mathcal{L}_{\mathrm{AR}}
=
-\frac{
\sum_{g\in\mathcal{G}} v_g
\sum_{j=1}^{L_g}
\log p_{\theta}
\left(
y_{t,j}^{g}
\mid
c_t, y_t^{<g}, y_{t,<j}^{g}
\right)
}{
\sum_{g\in\mathcal{G}} v_g L_g
}.
\label{eq:ar_loss}
\end{equation}
The autoregressive objective assigns equal weight to each valid
supervised action token, with unavailable action groups excluded by
$v_g$.

This joint training differs from $\Psi_0$~\cite{psi0}, which processes
the human and humanoid datasets in separate stages using distinct
action decoders and objectives. Our shared representation instead
trains one action-token model across both data sources.

\paragraph{Stage II: humanoid post-training}
We continue training the same autoregressive model on the SIMPLE
humanoid training set~\cite{simple}. These demonstrations align the
cross-embodiment representation with the target robot's observations,
actions, and interaction dynamics. This stage produces a
humanoid-aligned autoregressive checkpoint, which serves as the common
initialization for the subsequent stages and as the Holo-M AR baseline.

\paragraph{Stage III: grouped discrete diffusion fine-tuning}
We fine-tune the Stage-II checkpoint using the grouped
masked-diffusion objective in Eq.~\eqref{eq:diff_loss}. This stage
preserves the VLM backbone, canonical action vocabulary, and fixed
group order defined in Table~\ref{tab:canonical_action_space}. It
changes the action-generation process from token-level autoregression
to parallel de-masking within each group, while retaining causal
conditioning across groups. The resulting checkpoint is the Holo-M
generalist policy.

\paragraph{Stage IV: task-specific fine-tuning}
For specialist evaluation, we initialize the model from the Stage-III
checkpoint and fine-tune it on demonstrations from an individual
target task. We retain the same grouped discrete diffusion objective,
model architecture, and action vocabulary. This stage adapts the
generalist policy to the target environment and task dynamics without
introducing a separate action head. The resulting checkpoint is Holo-M specialist policy.

\section{Experiments}
\label{sec:experiments}

\subsection{Datasets and Evaluation Benchmark}
\label{sec:datasets_benchmark}

\subsubsection{Overview}
We follow the pre-training data configuration of
$\Psi_0$~\cite{psi0}, using EgoDex and Humanoid Everyday, and adopt
SIMPLE~\cite{simple} as the simulation benchmark. SIMPLE is designed
specifically for whole-body humanoid loco-manipulation, making it
better aligned with our setting than benchmarks for fixed-base or
wheeled manipulation.

SIMPLE provides three properties that are important for our
evaluation. First, the benchmark reports that
policy rankings in simulation well correlate with real-world performance. Second, its tasks, scenes and object assets cover fairly diverse loco-manipulation behaviors. Third, its three domain-randomization levels provide
controlled visual and spatial variations for evaluating both
generalist and specialist policies.

\subsubsection{Datasets}
For direct comparability with SIMPLE~\cite{simple}, we use the same three
data sources (Table~\ref{tab:data_benchmarks}): EgoDex and Humanoid
Everyday for autoregressive pre-training, followed by embodiment-specific
post-training on SIMPLE. EgoDex~\cite{egodex} contributes human-centric RGB images, 3D hand poses, and language annotations, while Humanoid Everyday
(HE)~\cite{humanoideveryday} contributes RGB observations and robot
actions from its Unitree G1 subset. We convert both datasets into the
canonical representation as described in
Sec.~\ref{sec:action_tokenizer}. The SIMPLE humanoid training
set~\cite{simple} provides embodiment-specific observations, states,
and actions for subsequent model adaptation. The complete training
schedule is described in Sec.~\ref{sec:training_pipeline}.

\subsubsection{Benchmarking}

\textbf{SIMPLE benchmark.}
Following the SIMPLE evaluation protocol~\cite{simple}, we evaluate all
policies on six representative tasks. These tasks collectively cover
rigid-object manipulation, non-prehensile interaction, articulated-object
manipulation, bimanual coordination, and mobile loco-manipulation.
Performance is determined by task-specific programmatic conditions, such
as lifting an object above a prescribed height, placing it within a target
region, or rotating an articulated joint beyond a specified angle.

\textbf{Evaluation protocol.}
We evaluate every method on the six tasks under the three SIMPLE
domain-randomization settings, denoted as Level~0,~1, and ~2~\cite{simple}. These settings vary distractors, visual
appearance and lighting, and the initial poses of the robot and
objects. We conduct 10 rollouts for each combination of method, task,
and randomization level, same as~\cite{simple}. A rollout is successful only if the
task-specific terminal condition is satisfied within the prescribed
control horizon. We follow the same observation space, language
instructions, initial-state distributions, and rollout budget for all
evaluated methods.

\begin{table}[t]
\centering
\caption{Datasets and benchmark used for training.}
\label{tab:data_benchmarks}
\adjustbox{max width=\textwidth}{
\begin{tabular}{lccccc}
\hline
Resource & Domain & Trajectories & Tasks & Modalities used & Role \\
\hline
EgoDex
& Human
& 829 h / 338k
& 194
& RGB, hand pose
& AR pre-training \\

Humanoid Everyday
& Humanoid
& 31 h / 10.3k
& 260
& RGB, action
& AR pre-training \\

SIMPLE
& Humanoid
& 1m
& 24
& RGB, state, action
& Post-training \\

\hline
\end{tabular}}
\end{table}

\subsection{Generalist Performance}
\label{sec:generalist}
We define a \emph{generalist} policy as a single checkpoint, post-trained
jointly on all 24 SIMPLE tasks~\cite{simpledata} with no
per-task fine-tuning, then evaluated without adaptation on the six
representative tasks used throughout this section -- in contrast to the
\emph{specialist} setting (next subsection), where a separate checkpoint is
fine-tuned per task. This tests whether joint training across a broad,
heterogeneous task distribution retains enough shared competence for each
task without task-specific adaptation -- the practically relevant regime
for deploying one policy across many tasks rather than maintaining a
bespoke checkpoint for each.

We compare Holo-M against the strongest generalist VLA/WAM baselines on
SIMPLE~\cite{simple}: $\Psi_0$~\cite{psi0}, $\pi_{0.5}$~\cite{pi05},
DreamZero~\cite{dreamzero}, and ACT~\cite{act}, all fine-tuned on the same
data as Holo-M for a consistent comparison. $\Psi_0$ fine-tuned only its
action head from the released VLM checkpoint; ACT trained from its ImageNet checkpoint;
$\pi_{0.5}$ initialized from \texttt{pi05\_droid} and only its action head is fine-tuned; and DreamZero
initialized from DreamZero-AgiBot and fine-tuned with LoRA after
converting the shared data to its native schema. All models kept their
original architectures and action-generation formulations.

We additionally report \textbf{Holo-M AR}, an ablation that decodes each
body part's tokens autoregressively instead of via the grouped discrete diffusion
scheme of Sec.~\ref{sec:grouped_diffusion}, sharing the same backbone,
tokenizer, and training data as Holo-M; comparing the two isolates
grouped discrete diffusion decoding's contribution. Table~\ref{table_generalist}
reports success under the same three domain-randomization levels (Level
0/1/2) and six tasks defined in~\cite{simple}; our policy uses 8 diffusion
steps.

\begin{table*}[t]
\centering
\caption{Generalist success rate (Level 0/1/2) on six SIMPLE
  loco-manipulation tasks.}
\label{table_generalist}
\adjustbox{max width=\textwidth}{
\begin{tabular}{lccccccc}
\toprule
Method & XMovePick & BendPick & Handover & Mobile P\&P & Grasp & XMoveBendPick & Overall \\
\midrule
$\Psi_0$~\cite{psi0} & \textbf{9/10/9} & 4/2/4 & \textbf{10/10/9} & 6/6/3 & \underline{8/7/6} & 4/4/3 & 114/180 \\
$\pi_{0.5}$~\cite{pi05} & 0/0/3 & 0/0/0 & 6/1/6 & 0/0/0 & 6/4/2 & 0/0/0 & 28/180 \\
DreamZero~\cite{dreamzero} & 0/0/0 & 0/0/0 & 7/7/6 & 0/0/0 & 7/5/6 & 0/0/0 & 38/180 \\
ACT~\cite{act} & 0/0/0 & \textbf{10/9/10} & 0/0/0 & \textbf{6/8/9} & 5/6/8 & 0/5/3 & 79/180 \\
\textbf{Holo-M AR (Ours)} & \underline{9/5/8} & \underline{10/8/8} & 9/6/8 & \underline{5/6/7} & \textbf{8/9/8} & \underline{7/8/9} & \underline{138/180} \\
\textbf{Holo-M (Ours)} & 9/7/3 & \underline{9/9/8} & \underline{7/7/10} & \textbf{8/9/6} & \textbf{8/9/8} & \textbf{9/8/9} & \textbf{143/180} \\
\bottomrule
\end{tabular}}
\end{table*}

Both Holo-M variants substantially outperform every baseline
(Table~\ref{table_generalist}), where bolded and underlined results are the best and second-best respectively. Holo-M reaches 143/180 overall and the
autoregressive ablation, Holo-M AR, reaches 138/180, both far ahead of
$\Psi_0$ (114/180) and the rest (under 80/180). That Holo-M AR alone beats
every baseline suggests our tokenizer and joint training recipe drive
most of this gain. Interestingly, grouped discrete diffusion decoding also achieves slightly higher aggregate success than the AR variant (143/180 vs. 138/180), while substantially reducing inference latency. Although this difference is modest, it suggests that parallel within-group decoding need not trade task performance for speed.

\subsection{Specialist Performance}
We define a \emph{specialist} policy as a checkpoint that is fine-tuned
individually for each of the six evaluation tasks, following the
single-task fine-tuning convention used for the baselines reported in the
$\Psi_0$~\cite{psi0} and SIMPLE~\cite{simple} papers. Our Holo-M specialists are
fine-tuned from the same generalist checkpoint, each using the corresponding single-task data.

Table~\ref{table_specialist} reports the full set of single-task
specialist baselines benchmarked in~\cite{simple}, including VLA models, world-action models
alongside Holo-M. Our Holo-M specialist attains the highest overall
success rate (163/180), ahead of the strongest baseline $\Psi_0$
(154/180). The
gap is largest on Mobile P\&P, the longest-horizon loco-manipulation task
among the six -- it requires the robot to navigate between tables while
carrying and placing an object, compounding locomotion and manipulation
over a much longer action horizon than the other five tasks. There, the
Holo-M specialist succeeds in 25 of 30 rollouts (8/9/8 across the three
randomization levels) versus 18 of 30 for the $\Psi_0$ specialist (7/5/6),
a relative improvement of nearly 40\%.

\begin{table*}[t]
\centering
\caption{Specialist success rate (Level 0/1/2) on six SIMPLE
  loco-manipulation tasks. Baseline numbers copied from~\cite{simple}.}
\label{table_specialist}
\adjustbox{max width=\textwidth}{
\begin{tabular}{lccccccc}
\toprule
Method & XMovePick & BendPick & Handover & Mobile P\&P & Grasp & XMoveBendPick & Overall \\
\midrule
$\Psi_0$ & 10/10/6 & \textbf{10/10/10} & 7/7/10 & 7/5/6 & \textbf{10/10/8} & \underline{10/9/9} & 154/180 \\
GR00T N1.6 & \underline{10/10/7} & 7/7/6 & 1/3/3 & 0/0/0 & 9/9/7 & 4/4/1 & 88/180 \\
$\pi_{0.5}$ & 7/5/1 & \underline{10/10/8} & 5/4/5 & 3/3/3 & \textbf{10/10/8} & 0/0/0 & 92/180 \\
InternVLA-M1 & 0/0/0 & 5/5/0 & 0/0/0 & 0/0/0 & 0/0/0 & 3/5/7 & 25/180 \\
H-RDT & 0/0/2 & 0/0/1 & 0/1/0 & 0/0/0 & 0/0/0 & 0/0/0 & 4/180 \\
DreamZero & \textbf{10/10/10} & 9/9/8 & 7/8/9 & 5/3/3 & 9/10/7 & 0/0/1 & 118/180 \\
EgoVLA & 0/1/2 & 7/5/8 & 0/4/3 & 0/0/0 & \underline{10/10/7} & 3/5/4 & 69/180 \\
DP & 3/3/2 & 10/8/6 & 3/2/4 & 4/0/0 & 8/9/8 & 0/0/0 & 70/180 \\
ACT & 10/10/5 & \underline{10/9/9} & 7/7/10 & 5/5/5 & \textbf{10/10/8} & \textbf{9/10/10} & 149/180 \\
\textbf{Holo-M AR (Ours)} & \underline{10/10/7} & \underline{9/10/9} & \underline{10/8/9} & \underline{8/8/8} & 9/10/7 & 8/7/10 & \underline{157/180} \\
\textbf{Holo-M (Ours)} & \textbf{10/10/10} & \underline{10/10/8} & \textbf{10/8/10} & \textbf{8/9/8} & \underline{9/9/9} & 8/8/9 & \textbf{163/180} \\
\bottomrule
\end{tabular}}
\end{table*}

\subsection{Data Scaling}
Table~\ref{table_data_scaling} ablates the effect of progressively adding
pre-training data sources, holding the backbone (Qwen3-VL~\cite{qwen3vl})
and the EEF~$\to$~body~$\to$~hand~$\to$~kinematics token ordering fixed.
\emph{No Pretraining} skips the autoregressive pretraining stage entirely
and post-trains the base VLM directly on the SIMPLE tasks;
\emph{HE} runs pretraining on Humanoid Everyday~\cite{humanoideveryday}
alone; \emph{EgoDex+HE} is our full pretraining recipe. We report
both an open-loop metric (mean absolute error, MAE, between predicted and
ground-truth detokenized actions) and
closed-loop success rate on SIMPLE~\cite{simple}, so as to separate how
well the model imitates the training distribution from how well that
translates into task completion under closed-loop execution.

\begin{table}[t]
\centering
\caption{Effect of scaling pre-training data on Holo-M AR.}
\label{table_data_scaling}
\begin{center}
\begin{tabular}{lcc}
\toprule
Pretraining Data & Open-loop MAE $\downarrow$ & Closed-loop Success Rate $\uparrow$ \\
\midrule
No Pretraining & 0.022 & 53/180 \\
HE & 0.020 & 119/180 \\
EgoDex + HE & 0.019 & 138/180 \\
\bottomrule
\end{tabular}
\end{center}
\end{table}

Pretraining on
Humanoid Everyday alone doubles closed-loop success (53/180 to 119/180). Adding EgoDex on top of HE further improves success
rate to 138/180. This indicates that once whole-body
humanoid grounding is established from robot data, additional egocentric
human data sharpens the model's closed-loop robustness. Overall, pretraining data alone is
responsible for a 2.6$\times$ increase in success rate (53/180 to
138/180),
suggesting that the heterogeneous pretraining sources are crucial for scaling Holo-M's performance.

\subsection{Real-World Deployment}

\textbf{Hardware.}
We deploy on a similar platform as~\cite{psi0}: a Unitree G1-comp (competition
edition) with two Dex3 hands and a top-mounted RealSense camera. The
policy consumes the camera at $640\times360$, 30\,Hz, the 31 upper-body
joint angles (14 hand, 14 arm, waist roll/pitch/yaw) binned into 256 state
tokens each, and the task instruction; lower-body joints, base velocity,
    height and IMU are not inputs. It outputs one second of action and runs on one RTX~5090 over
WebSocket. We reuse the Psi-0 real-robot stack -- client, observation/action
protocol and whole-body controller -- replacing only the policy server, so
any performance difference is attributable to the policy and chunk
scheduling alone.

\begin{figure}[t]
\centering
\includegraphics[width=0.8\columnwidth]{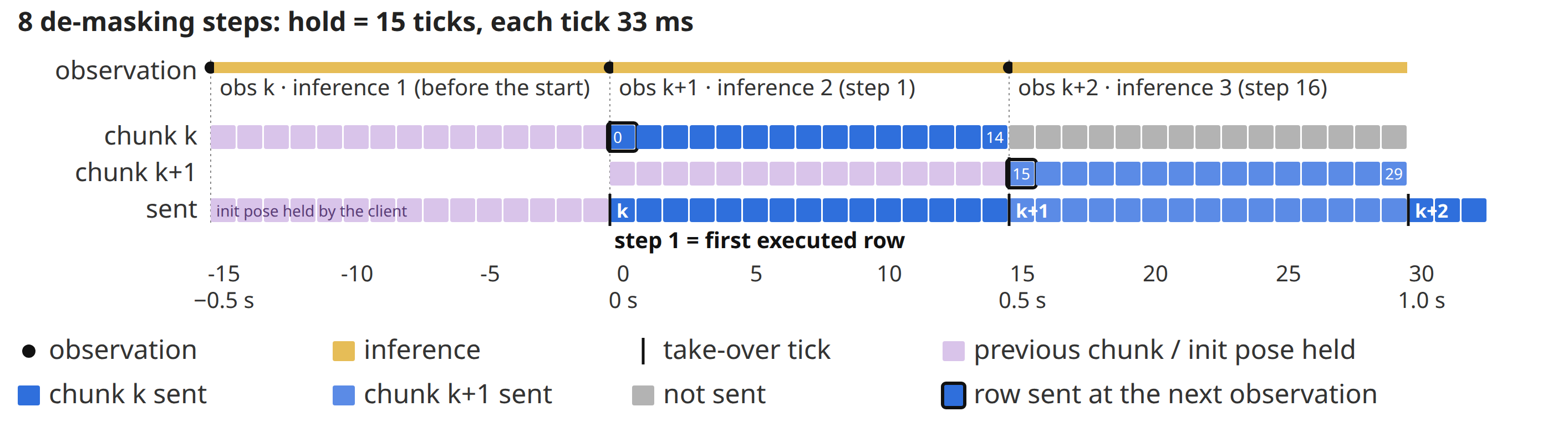}
\caption{Our fixed schedule along the tick axis using 8 de-masking steps as example: observation is taken every 500ms, which the model consumes to infer 1-second action chunk. During model inference, the action steps from the previous chunk are executed.}
\label{fig:fixsched}
\end{figure}

\textbf{Control loop.}
\emph{Our implementation: fixed schedule (Fig.~\ref{fig:fixsched}).} While
the model infers a new chunk, the robot keeps executing the previous
chunk. A new observation is captured on a fixed
cadence, once every 15 ticks (500\,ms), within which the model
finishes its inference inside a fixed hold budget and outputs a full
30-timestep (one-second) action chunk. The very first chunk is issued
before the robot moves and is never held, so it takes over immediately
from its own first predicted step. Every later chunk, however,
discards its own leading predicted steps that fall within the hold --
since the previous chunk is still executing those -- and instead takes
over starting from the step right after the hold, continuing for the
remaining 15 steps until the next chunk is ready. Because every
observation is captured on the same fixed 15-tick grid, every later
chunk always takes over at the same fixed point in that grid. In our experiment, the hold is
233, 333 and 500\,ms for 2, 4 and 8 de-masking steps respectively; if
inference exceeds the hold, the robot returns to
its previous pose. Table~\ref{tab:steps} gives the measured latencies for different de-masking steps.

\begin{table}[t]
\centering
\caption{Real-world inference latency per 30-timestep chunk.}
\label{tab:steps}
\begin{tabular}{c c}
\toprule
De-masking steps & Inference latency, mean $\pm$ std (ms) $\downarrow$ \\
\midrule
2 & $175.8 \pm 11.5$ \\
4 & $275.7 \pm 9.3$  \\
8 & $476.8 \pm 16.9$ \\
\bottomrule
\end{tabular}
\end{table}

\textbf{Deployment.} We show the deployment on two typical tasks, Tabletop Grasp and Move Pick. Each deployed checkpoint is a specialist
fine-tuned from the same generalist (described in Sec.~\ref{sec:generalist}) checkpoint, on a teleoperated dataset of 100
episodes, collected in the deployment scene for that task, using the same data collection pipeline as~\cite{psi0}. Table~\ref{tab:deploy_success}
reports real-robot success rate over 10 rollouts per task. Videos of the
G1-comp successfully carrying out Tabletop Grasp and Move Pick tasks
under this deployment stack are available on our project page:
\url{https://horizonrobotics.github.io/gail/Holo-M/}.

\begin{table}[t]
\centering
\caption{Real-robot success rate over 10 rollouts per task.}
\label{tab:deploy_success}
\begin{tabular}{lc}
\toprule
Task & Success Rate \\
\midrule
Tabletop Grasp & 10/10 \\
Move Pick & 8/10 \\
\bottomrule
\end{tabular}
\end{table}

\section{Conclusions}
\label{sec:conclusion}

We have presented Holo-M, to intrinsically exploit the language
model by extending its vocabulary with action tokens. On the SIMPLE benchmark~\cite{simple}, Holo-M outperforms all
the baseline models by significant margins.
This work is our first version, i.e. Holo-M v1, and several directions remain open for future
models built on this framework. First, because Holo-M places action tokens
in the same vocabulary as language, its capacity for long-horizon,
language-specified loco-manipulation should scale directly with continued
improvements in the backbone's reasoning ability -- for instance, by
interleaving explicit reasoning tokens, such as subtask decomposition or
failure recovery, with action tokens, rather than requiring a separately
scaled action expert. Second, our unified per-body-part tokenizer is
designed to absorb heterogeneous data sources beyond what we use here; a
natural next step is to scale training with substantially larger amounts of
ego-centric human video, which is far more abundant than teleoperated robot
data and could further improve generalization across tasks and embodiments.
Finally, Holo-M v1 is trained and evaluated with a decoupled WBC~\cite{decoupledwbc}.
Our future versions will remap the
tokenizer's detokenization to other controllers' e.g. HoloMotion~\cite{holomotion} and SONIC~\cite{sonic},
and leaving the underlying tokenization, attentions between modalities, and de-masking decoding
scheme unchanged.

\section*{Acknowledgments}
Figs.~\ref{fig:overview}, \ref{fig:training_pipeline}, and
\ref{fig:fixsched} were rendered with the assistance of Claude
(Anthropic) and Codex (OpenAI); the authors designed the content and
structure, and verified the rendered figure.


\begin{thebibliography}{99}

\bibitem{rt2} A. Brohan, N. Brown, J. Carbajal, \emph{et al.}, ``RT-2:
  Vision-language-action models transfer web knowledge to robotic control,''
  in \emph{Proc. Conf. on Robot Learning (CoRL)}, 2023.

\bibitem{openvla} M. J. Kim, K. Pertsch, S. Karamcheti, \emph{et al.},
  ``OpenVLA: An open-source vision-language-action model,'' in \emph{Proc.
  Conf. on Robot Learning (CoRL)}, 2024.

\bibitem{discretediffusionvla} Z. Liang, Y. Li, T. Yang, \emph{et al.},
  ``Discrete diffusion VLA: Bringing discrete diffusion to action decoding
  in vision-language-action policies,'' in \emph{Proc. Int. Conf. on
  Machine Learning (ICML)}, 2026.

\bibitem{knowledgeinsulation} D. Driess, J. T. Springenberg, B. Ichter,
  \emph{et al.}, ``Knowledge insulating vision-language-action models:
  Train fast, run fast, generalize better,'' in \emph{Proc. Advances in
  Neural Information Processing Systems (NeurIPS)}, 2025.

\bibitem{psi0} S. Wei, H. Jing, B. Li, \emph{et al.}, ``$\Psi_0$: An open
  foundation model towards universal humanoid loco-manipulation,'' in
  \emph{Proc. Robotics: Science and Systems (RSS)}, 2026.

\bibitem{simple} S. Wei, Z. Ni, J. Liu, \emph{et al.}, ``SIMPLE:
  Simulation-based policy learning and evaluation for humanoid
  loco-manipulation,'' \emph{arXiv preprint arXiv:2606.08278}, 2026.

\bibitem{simpledata} USC-PSI-Lab, ``psi-data: SIMPLE benchmark dataset,''
  Hugging Face dataset repository, 2026. [Online]. Available:
  \url{https://huggingface.co/datasets/USC-PSI-Lab/psi-data/tree/a7d4cb67506566aae82ea9120516547a10057b10/simple}
  
\bibitem{g05} Y. Liu, Z. Dong, B. Ye, \emph{et al.}, ``G0.5: One
  Autoregressive Stream for Robot Reasoning and Action,'' \emph{arXiv
  preprint arXiv:2608.11739}, 2026.

\bibitem{wholebodyvla} H. Jiang, J. Chen, Q. Bu, \emph{et al.},
  ``WholeBodyVLA: Towards unified latent VLA for whole-body
  loco-manipulation control,'' in \emph{Proc. Int. Conf. on Learning
  Representations (ICLR)}, 2026.

\bibitem{pi0} K. Black, N. Brown, D. Driess, \emph{et al.}, ``$\pi_0$: A
  vision-language-action flow model for general robot control,'' \emph{arXiv
  preprint arXiv:2410.24164}, 2024.

\bibitem{fast} K. Pertsch, K. Stachowicz, B. Ichter, \emph{et al.}, ``FAST:
  Efficient action tokenization for vision-language-action models,'' in
  \emph{Proc. Robotics: Science and Systems (RSS)}, 2025.

\bibitem{faster} Y. Liu, S. Zhang, Z. Dong, \emph{et al.}, ``FASTer: Toward
  efficient autoregressive vision language action modeling via neural
  action tokenization,'' \emph{arXiv preprint arXiv:2512.04952}, 2025.

\bibitem{holomotion} M. Chen, K. Wang, B. Zhang, \emph{et al.},
  ``HoloMotion-1 technical report,'' \emph{arXiv preprint
  arXiv:2605.15336}, 2026.

\bibitem{sonic} Z. Luo, Y. Yuan, T. Wang, \emph{et al.}, ``SONIC:
  Supersizing motion tracking for natural humanoid whole-body control,''
  \emph{Science Robotics}, 2026.

\bibitem{decoupledwbc} NVIDIA, ``GR00T whole-body control,'' GitHub
  repository, 2025. [Online]. Available:
  \url{https://github.com/NVlabs/GR00T-WholeBodyControl}

\bibitem{maskgit} H. Chang, H. Zhang, L. Jiang, \emph{et al.}, ``MaskGIT:
  Masked generative image transformer,'' in \emph{Proc. IEEE/CVF Conf. on
  Computer Vision and Pattern Recognition (CVPR)}, 2022.


\bibitem{egodex}
R. Hoque, P. Huang, D. J. Yoon, \emph{et al.}, ``EgoDex: Learning
dexterous manipulation from large-scale egocentric video,'' in
\emph{Proc. Int. Conf. on Learning Representations (ICLR)}, 2026.

\bibitem{humanoideveryday}
Z. Zhao, H. Jing, X. Liu, \emph{et al.}, ``Humanoid Everyday: A
comprehensive robotic dataset for open-world humanoid manipulation,''
in \emph{Proc. IEEE Int. Conf. on Robotics and Automation (ICRA)}, 2026.

\bibitem{qwen3vl}
S. Bai, Y. Cai, R. Chen, \emph{et al.}, ``Qwen3-VL Technical Report,''
  \emph{arXiv preprint arXiv:2511.21631}, 2025.

\bibitem{pi05} K. Black, N. Brown, J. Darpinian, \emph{et al.},
  ``$\pi_{0.5}$: A vision-language-action model with open-world
  generalization,'' in \emph{Proc. Conf. on Robot Learning (CoRL)}, 2025.

\bibitem{dreamzero} S. Ye, Y. Ge, K. Zheng, \emph{et al.}, ``World action
  models are zero-shot policies,'' \emph{arXiv preprint arXiv:2602.15922},
  2026.

\bibitem{act} T. Z. Zhao, V. Kumar, S. Levine, and C. Finn, ``Learning
  fine-grained bimanual manipulation with low-cost hardware,'' in
  \emph{Proc. Robotics: Science and Systems (RSS)}, 2023.

\bibitem{diffusionvl}
L. Zeng, J. Yao, B. Liao, \emph{et al.},
``DiffusionVL: Translating any autoregressive models into diffusion
vision language models,'' in \emph{Proc. European Conf. on Computer
Vision (ECCV)}, 2026.

\end{thebibliography}
\end{document}